\documentclass[journal]{IEEEtran}

\IEEEoverridecommandlockouts
\usepackage{graphicx}
\def\BibTeX{{\rm B\kern-.05em{\sc i\kern-.025em b}\kern-.08em
    T\kern-.1667em\lower.7ex\hbox{E}\kern-.125emX}}
\usepackage[font=small,labelfont=bf,tableposition=top]{caption}

\usepackage{blindtext}
\usepackage{caption}

\usepackage{times}
\usepackage{epsfig}
\usepackage{graphicx}
\usepackage{amsmath}
\usepackage{amssymb}
\usepackage{color}
\usepackage{booktabs}
\usepackage{bbm}
\usepackage{bm}
\usepackage{multirow}
\usepackage{indentfirst}
\usepackage{hhline}
\usepackage{array}
\usepackage{arydshln}
\newcolumntype{C}[1]{>{\centering}p{#1}}
\usepackage{makecell}
\usepackage{pifont}
\usepackage{caption}
\usepackage{graphicx}
\usepackage{float}
\usepackage{subcaption}
\usepackage{mathrsfs}
\usepackage[table]{xcolor}
\definecolor{rblue}{rgb}{0,0.5,1}
\usepackage[breaklinks,colorlinks]{hyperref}
\hypersetup{colorlinks, citecolor=rblue}

\begin{document}
\title{SSD-MonoDETR:\\Supervised Scale-aware Deformable Transformer for Monocular 3D Object Detection}

\author{Xuan He, Fan Yang, Kailun Yang\IEEEauthorrefmark{2}, Jiacheng Lin, Haolong Fu, Meng Wang,~\IEEEmembership{Fellow,~IEEE},\\Jin Yuan\IEEEauthorrefmark{1}\IEEEauthorrefmark{2}, and Zhiyong Li\IEEEauthorrefmark{1}%
\thanks{This work was supported in part by the National Natural Science Foundation of China (No.61976086, No.U21A20518, and No.62272157), in part by the Natural Science Foundation of Changsha (No. kq2202177), and in part by Hangzhou SurImage Technology Company Ltd. \textit{(Corresponding authors: Jin Yuan and Zhiyong Li.)}}
\thanks{X. He, F. Yang, J. Lin, H. Fu, J. Yuan, and Z. Li are with the College of Computer Science and Electronic Engineering, Hunan University, Changsha 410082, China.}
\thanks{K. Yang and Z. Li are with the School of Robotics, Hunan University, China 410012, China.}
\thanks{K. Yang and Z. Li are also with the National Engineering Research Center of Robot Visual Perception and Control Technology, Hunan University, Changsha 410082, China.}%
\thanks{M. Wang is with the School of Computer Science, Hefei University of Technology, Hefei 230009, China.}
\thanks{\IEEEauthorrefmark{1}Corresponding authors: Jin Yuan and Zhiyong Li. (E-mail: yuanjin@hnu.edu.cn, zhiyong.li@hnu.edu.cn.)}
\thanks{\IEEEauthorrefmark{2}Equal advising.}
}

% The paper headers
\markboth{IEEE Transactions on Intelligent Vehicles, September~2023}%
{He \MakeLowercase{\textit{et al.}}: SSD-MonoDETR}

\maketitle

\IEEEpeerreviewmaketitle

\begin{abstract}
Transformer-based methods have demonstrated superior performance for monocular 3D object detection recently, which aims at predicting 3D attributes from a single 2D image. Most existing transformer-based methods leverage both visual and depth representations to explore valuable query points on objects, and the quality of the learned query points has a great impact on detection accuracy. Unfortunately, existing unsupervised attention mechanisms in transformers are prone to generate low-quality query features due to inaccurate receptive fields, especially on hard objects. To tackle this problem, this paper proposes a novel ``Supervised Scale-aware Deformable Attention'' (SSDA) for monocular 3D object detection. Specifically, SSDA presets several masks with different scales and utilizes depth and visual features to adaptively learn a scale-aware filter for object query augmentation. Imposing the scale awareness, SSDA could well predict the accurate receptive field of an object query to support robust query feature generation. Aside from this, SSDA is assigned with a Weighted Scale Matching (WSM) loss to supervise scale prediction, which presents more confident results as compared to the unsupervised attention mechanisms. Extensive experiments on the KITTI and Waymo Open datasets demonstrate that SSDA significantly improves the detection accuracy, especially on moderate and hard objects, yielding state-of-the-art performance as compared to the existing approaches. Our code will be made publicly available at \url{https://github.com/mikasa3lili/SSD-MonoDETR}.
\end{abstract}


%%%%%%%%% BODY TEXT
\section{Introduction}\label{sec:intro}

\IEEEPARstart{T}{he} recent progresses in 3D-related studies have tremendously promoted their wide applications in multiple object tracking~\cite{wang2023interactive, guo20223d}, depth and ego-motion estimations~\cite{liu2022self, wang20223d}, and object detection for autonomous driving~\cite{wang2022performance,wang2023multi} or indoor robotics~\cite{song2015sun, dai2017scannet}.
Benefiting from high-performance hardware, 3D object detection methods based on LiDAR points~\cite{zhou2018voxelnet, sheng2021improving, xu2022behind, meng2023hydro} or binocular images~\cite{chen20173d, li2019stereo, liu2021yolostereo3d} have achieved promising performance, but still suffer from high hardware costs. Comparatively, monocular 3D object detection~\cite{brazil2019m3d, weng2019monocular, liu2020smoke}, which predicts 3D attributes from a single image, could greatly save computation- and equipment costs and thus has attracted increasingly more research attention.\\
\begin{figure}[t]
	\centering
	\includegraphics[scale=0.3]{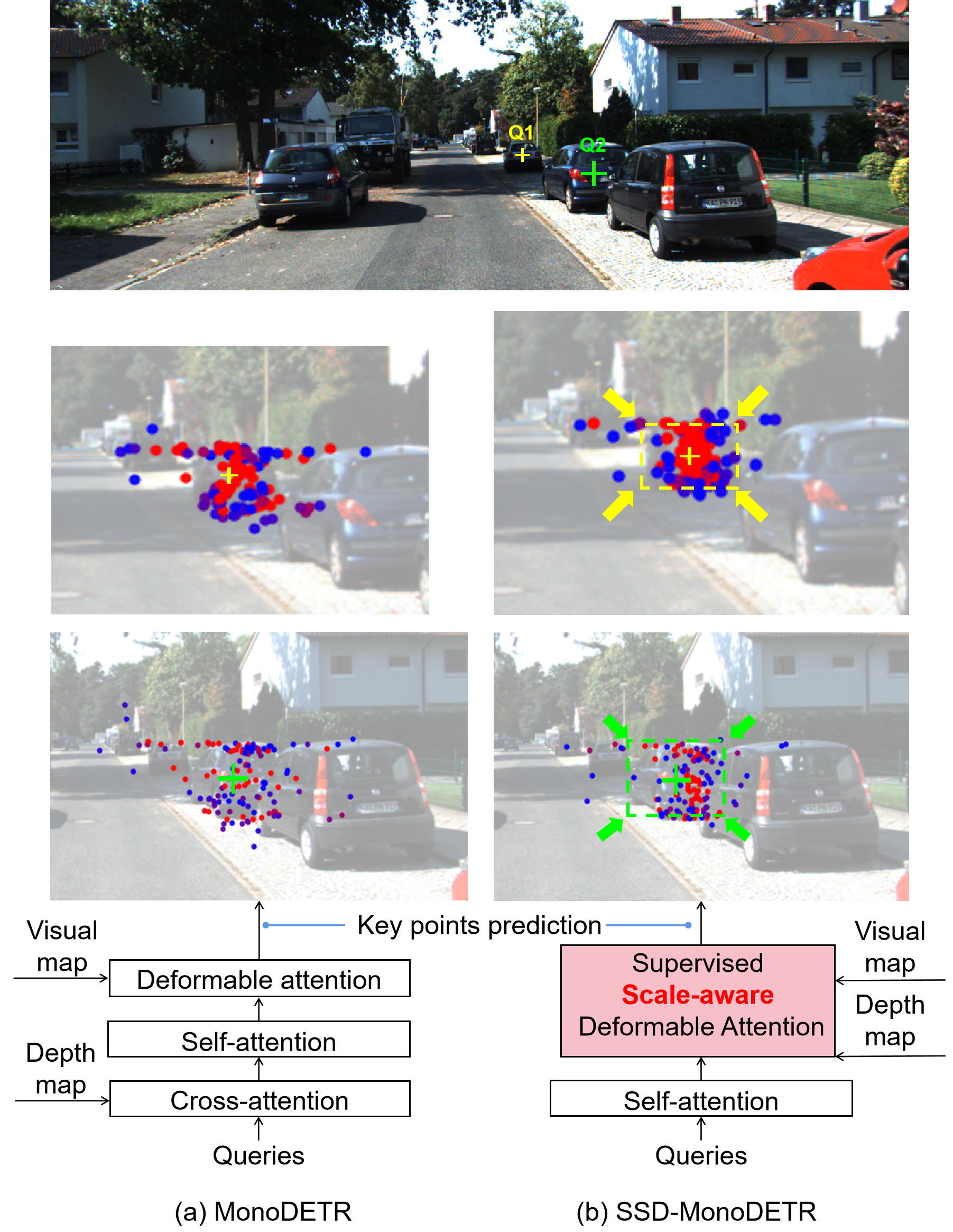}
	\caption{Two examples to visualize the predicted key points (cycle) on object queries (cross) by (a) MonoDETR and (b) SSD-MonoDETR, where the red indicates the higher attention weight. It is obvious that SSDA could generate key points with higher quality by using the scale-aware mechanism.}
	\label{intro}
\end{figure}
\indent Most existing monocular 3D detection methods revise the traditional 2D object detectors \cite{lin2017focal, ren2015faster, tian2019fcos}, which either require adjusting abundant manual parameters like anchors, region proposal and NMS~\cite{brazil2019m3d, brazil2020kinematic, ding2020learning, shi2021geometry}, or over-depend on exploring the geometric relationship between 2D and 3D objects \cite{ku2019monocular, wang2022probabilistic, su2023opa}. Comparatively, newly proposed transformer-based detectors like MonoDETR \cite{zhang2022monodetr} could avoid tedious parameter settings, and have demonstrated superior performance for monocular 3D object detection \cite{huang2022monodtr, wu2022dst3d, zhang2022monodetr}. Given an input image with random query points, MonoDETR \cite{zhang2022monodetr} first adopts several attention operations on visual and depth maps (see Figure \ref{intro} (a)) to search for relative key points for each query, and then integrates the features from these key points to predict 3D attributes. Technically, although deformable attention could find the amount of relative key points for a query, still suffers from the serious noisy point problem, especially on hard objects, where many key points deviate to background or irrelative objects (see Figure~\ref{intro}~(a)). 
This problem stems from the inherent mechanism of deformable attention, which is fed by a set of randomly initialized queries, and needs to adaptively explore relative key points for queries without the ability to estimate their receptive fields. Without the assistance of a precise receptive field, the predicted key points for a query tend to drift to other similar objects, generating noisy key points outside the object. As a result, the feature aggregation from noisy points would significantly affect 3D attribute prediction.\\
\indent To alleviate this problem, this paper proposes a Supervised Scale-aware Deformable Transformer for monocular 3D object detection (SSD-MonoDETR). Different from MonoDETR, SSD-MonoDETR reduces the depth cross-attention layer and introduces a novel Supervised Scale-aware Deformable Attention (SSDA) layer to emphasize key point prediction for object queries (see Figure \ref{intro}~(b)). Specifically, given a query, SSDA first presets several masks with different scales to extract multi-scale local features for the query from the input visual map, as well as predicts the scale probability distribution of the query from the depth map. On this basis, SSDA then adopts a lightweight adaptive layer to learn the receptive field of the query, which could offer valuable scale information for keypoint prediction. 
To guide the learning of SSDA, we design a Weighted Scale Matching (WSM) loss to extract the scale probability distribution output by the SSDA layer and use the scale ground truth to supervise it. As a result, the inner parameter updating of deformable attention could receive direct supervision, thus owning the ability to estimate the receptive field of a query to guide key point generation as compared to the original deformable attention. Benefiting from this, the noisy key point generation could be well alleviated, and more accurate query features are extracted for 3D attribute prediction (see Figure~\ref{intro}~(b)).
We conduct extensive experiments on the KITTI and Waymo Open datasets, and the experimental results demonstrate the effectiveness of SSDA especially on moderate and hard objects, yielding state-of-the-art performance as compared to the existing approaches.

At a glance, this work yields the following contributions:
\begin{enumerate}
	\item We propose a Supervised Scale-aware Deformable Attention (SSDA) mechanism to improve the quality of the learned object queries in transformers. Compared to deformable attention, SSDA could better predict the receptive field of an object query to support accurate key point generation, yielding high-quality query features for 3D attribute prediction.
	\item We design a Weighted Scale Matching (WSM) loss on SSDA to directly supervise the scale learning of queries without extra labeling costs, which is more effective as compared to the existing unsupervised attention mechanisms in transformers.
	\item An extensive set of experimental results on the KITTI and Waymo Open datasets demonstrates the leading performance of our approach as compared to the state-of-the-art approaches in moderate and hard subsets. Furthermore, the near real-time inference time indicates the high applicability of our method in real cases.
\end{enumerate}

In the rest of this paper, we first introduce the works related to our research in Section \ref{relat}, and then we elaborate on the details of our framework in Section \ref{method}. Finally, we conduct a comprehensive variety of comparative and evaluation experiments to verify the effectiveness of the proposed method in Section \ref{exp}, and give the conclusion, limitation, and future perspective of our work in Section \ref{conclusions}.
%-------------------------------------------------------------------------
\begin{figure*}[t!]
	\centering
	\includegraphics[scale=0.48]{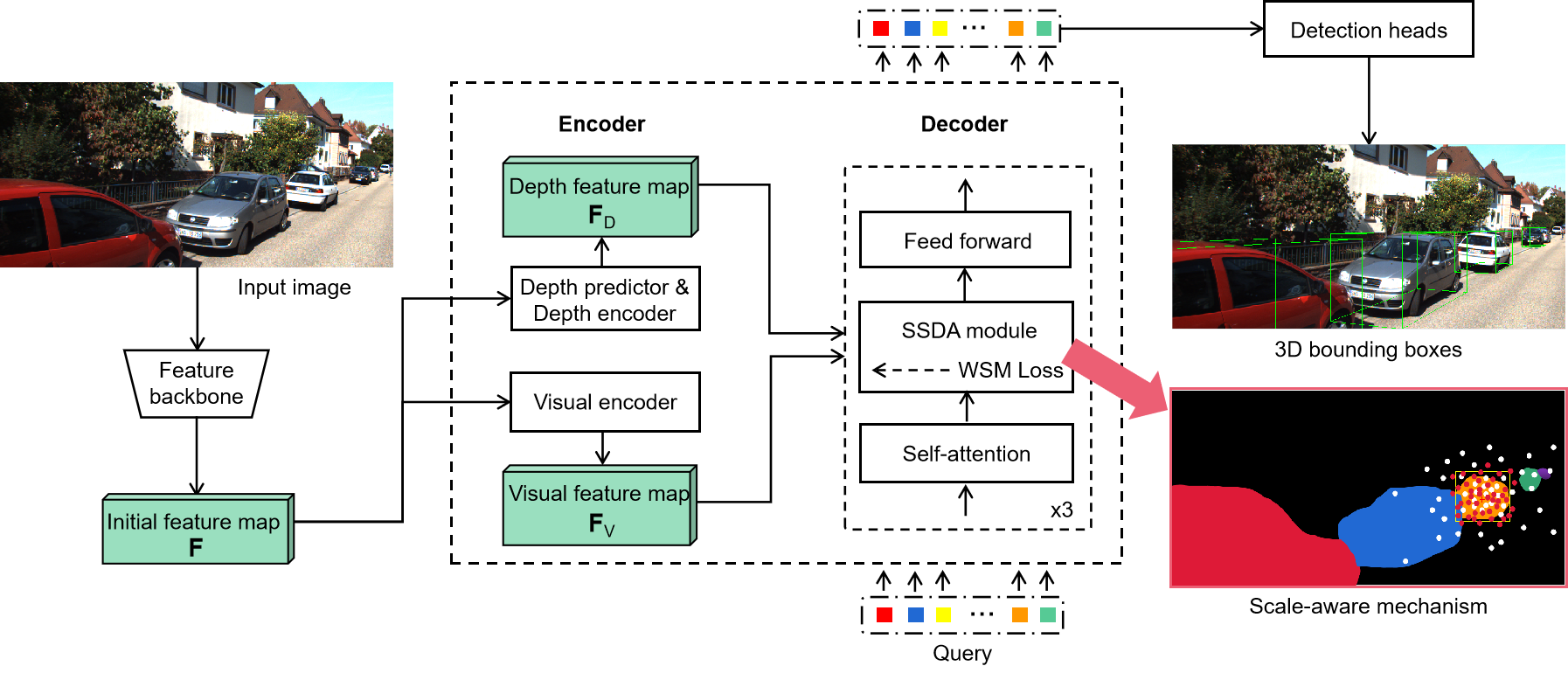}
	\caption{The architecture of SSD-MonoDETR, where SSDA with a WSM loss is introduced to localize the scope of each object query, yielding a more accurate query receptive field (the red points) as compared to MonoDETR (the white points).}
	\label{pipeline}
\end{figure*}
\section{Related Work} \label{relat}
Monocular 3D object detection aims to estimate the 3D attributes of objects from a single image, where depth estimation is an ill-pose problem thus especially difficult. To boost the performance, many approaches explore the assistance of external resources like depth maps~\cite{ding2020learning, ma2020rethinking, park2021pseudo, ma2019accurate} or LiDAR~\cite{huang2022monodtr, reading2021categorical, chen2021monorun}. Although effective, they inevitably bring extra costs to data collection and calculation. To this end, image-only methods without extra data have attracted increasing attention.

\textbf{Image-only monocular 3D object detection:}
Most existing image-only monocular 3D object detection approaches \cite{mousavian20173d, wang2021fcos3d, lu2021geometry, zhou2023bridging} revise their pipelines based on 2D object detectors \cite{chen2016monocular, chen20153d, simonelli2019disentangling}. To improve performance, the geometry prior or key point prediction are the common strategies. 
For the methods that use geometric prior, OFT-Net~\cite{roddick2018orthographic} proposes an orthographic feature transform to transcend the limitations of the image domain by converting image-based features into a 3D orthographic space. 
MonoRCNN~\cite{shi2021geometry} proposes a geometry-based distance decomposition to factor the object distance into more stable physical and projected 2D heights.
Based on MonoRCNN, MonoRCNN++~\cite{shi2023multivariate} further models the joint probability distribution of the physical height and visual height.
Monopair~\cite{chen2020monopair} further explores the spatial relationship between pairs of objects to augment the 3D location.
MonoJSG~\cite{lian2022monojsg} utilizes pixel-level geometric constraints to progressively refine the depth estimation. 
Another stream of methods first predicts the key points of the 3D bounding box and regards it as an auxiliary task.
For example, RTM3D~\cite{li2020rtm3d} predicts nine key points of a 3D bounding box to explore the 2D-3D geometric relationship to recover the 3D attributes.
SMOKE~\cite{liu2020smoke} is built based on the CenterNet~\cite{zhou2019objects}, which treats the objects as points and combines the key points estimation with 3D attributes regression.
MonoDLE~\cite{ma2021delving} revisits the issue of misalignment between the center of the 2D bounding box and the projected center of the 3D object, thus proposing to directly detect the projected 3D center.
MonoFlex~\cite{zhang2021objects} addresses the prediction of long-tail truncated objects by decoupling the edge of the feature map, whose backbone utilizes three perspective-projection-based and one direct-regression-based depth estimators. 
Compared to MonoFlex, PDR~\cite{sheng2023pdr} only requires one perspective-projection-based estimator to regress depth to realize a lighter architecture but better performance.
With complicated designs, the above methods have achieved performance improvements, but there is still room for breakthroughs in accuracy and speed. 

\textbf{Transformer-based object detection:}
Transformer~\cite{vaswani2017attention} was initially introduced in sequential modeling and has made significant advancements in the area of natural language processing (NLP).
For object detection, DETR~\cite{carion2020end} first designs a novel pipeline based on the successful self-attention mechanism in transformers and abandons the complex manual settings in traditional 2D detectors.
Based on DETR, there are many works \cite{zheng2020end, gao2021fast, dai2021up} that strive to make further improvements. For example, Anchor DETR~\cite{wang2022anchor} designs an anchor-based object query thus the object queries could focus on the objects near the anchor points. \cite{sun2021rethinking} proposes a feature of interest selection mechanism to tackle the slow convergence of DETR caused by the Hungarian loss and the cross-attention mechanism. Conditional DETR~\cite{meng2021conditional} learns a conditional spatial query for the decoder, which could shrink the spatial range for queries and thus realize faster training. To accelerate the training process as well as improve the performance on small objects, Deformable DETR~\cite{zhu2020deformable} proposes a novel deformable attention module where the queries only pay attention to a small set of key sampling points around themselves. 

\textbf{Transformer-based monocular 3D object detection:}
Inspired by the successful applications of transformers in 2D object detection~\cite{zhen2022towards, dai2022ao2, sun2022multi, liu2023continual}, recent researchers have turned to paying increasing attention to transformers for monocular 3D object detection, which could save cumbersome post-processing such as non-maximum suppression (NMS). For instance, MonoDTR~\cite{huang2022monodtr} introduces LiDAR point clouds as auxiliary supervision of its transformer pipeline and utilizes the learned depth features as the input query of a decoder. Without any extra data, DST3D proposes a novel structure~\cite{wu2022dst3d} combining Swin Transformer~\cite{liu2021swin} with deep layer aggregation~\cite{yu2018deep} to realize 3D object detection. MonoPGC~\cite{wu2023monopgc} proposes a depth-space-aware transformer and a depth-gradient positional encoding to combine 3D space positions with depth-aware features. MonoDETR~\cite{zhang2022monodetr} designs a depth-guided decoder, which utilizes a depth cross-attention layer to extract the global depth cues and a deformable attention layer to aggregate local visual features for queries. Thanks to the creative depth-guided decoder, MonoDETR achieves competitive performance.

Differently, this paper focuses on improving the quality of object queries in MonoDETR by imposing scale awareness for monocular 3D object detection. We newly design a Supervised Scale-aware Deformable Attention (SSDA) to replace deformable attention in MonoDETR, which could predict the receptive field of an object query to better support accurate feature generation for it.

\begin{figure*}[t!]
	\centering
	\includegraphics[scale=0.48]{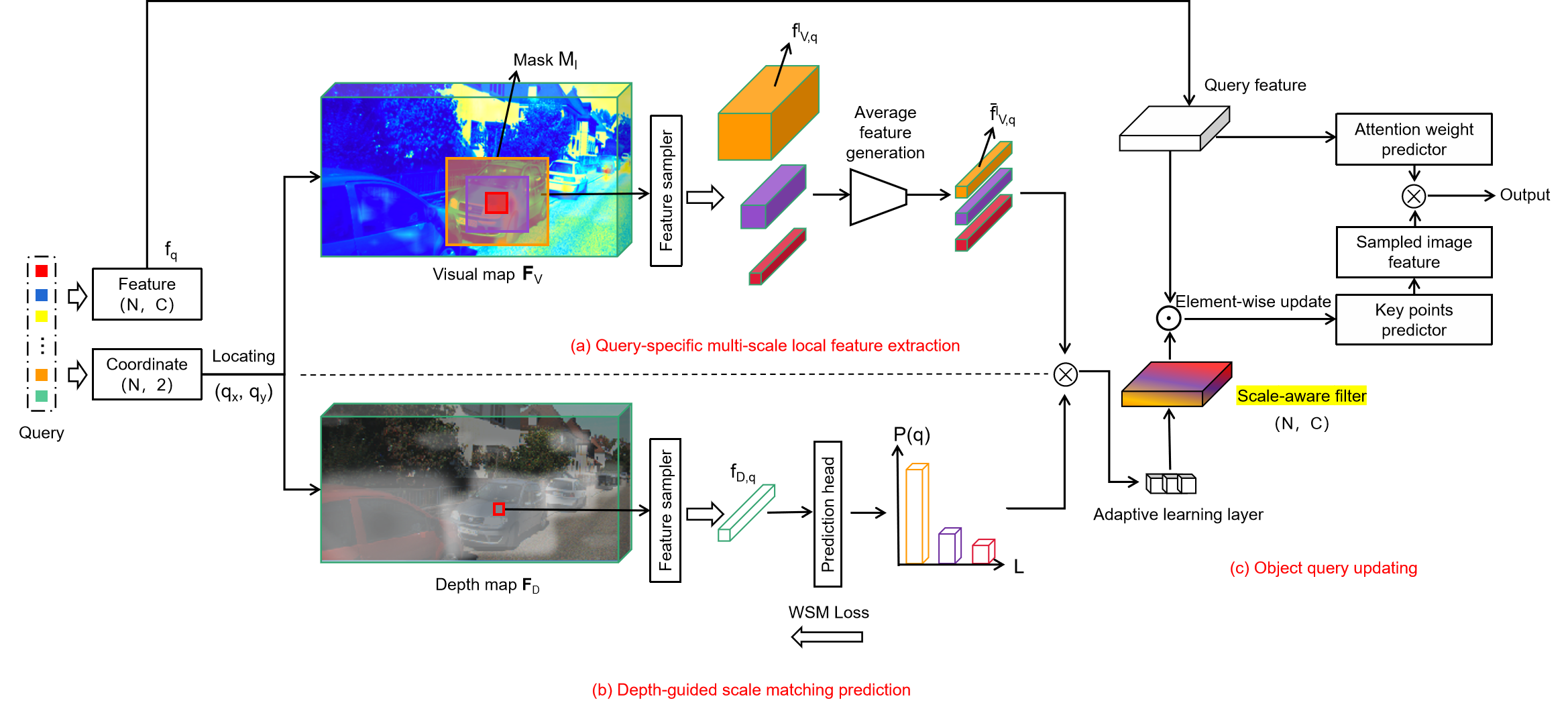}
	\caption{The detailed implementation of SSDA, is composed of three steps: (a) query-specific multi-scale local feature extraction, (b) depth-guided scale matching prediction, and (c) object query updating.}
	\label{SSDA}
\end{figure*}
\section{Method} \label{method}
\subsection{Architecture Overview}
Figure \ref{pipeline} demonstrates the architecture of the proposed Supervised Scale-aware Deformable Transformer for monocular 3D object detection (SSD-MonoDETR). Specifically, SSD-MonoDETR first adopts a feature backbone to generate the initial feature map for a given image. On this basis, the visual and depth encoders are employed to extract global visual and depth representations respectively, which are then passed to a transformer decoder to generate valuable queries for object prediction by detection heads. Compared to MonoDETR,  SSD-MonoDETR introduces a Supervised Scale-aware Deformable Attention (SSDA) module to integrate vision and depth information on each object query with a scale awareness, yielding more accurate receptive field as well as better multi-scale local feature representations for object queries. Moreover, we design a Weighted Scale Matching (WSM) loss for SSDA to impose scale awareness on object queries. As a result, our approach could better localize the scope of a query for accurate feature extraction.

\subsection{Supervised Scale-aware Deformable Attention}\label{DVMG}
SSDA belongs to the transformer decoder in SSD-MonoDETR and receives the visual embedding $\mathbf{F}_V$, the depth embedding $\mathbf{F}_D$, and a set of learnable queries $\mathbb{Q}$ from the last self-attention layer. Specifically, SSDA first extracts multi-scale local visual features for each query in $\mathbb{Q}$ (see Section~\ref{QMF}), as well as predicts the corresponding depth-guided scale matching probabilities for different scales (see Section~\ref{DMS}). Then, SSDA integrates multi-scale local features and scale-matching probabilities by using an adaptively learning layer to form a scale-aware filter, which is used for key point prediction and query updating (see Section~\ref{update}).

\subsubsection{Query-specific Multi-scale Local Feature Extraction}\label{QMF}
As Figure \ref{SSDA} shows, given the global visual representation $\mathbf{F}_V \in \mathbbm{R}^{\frac{H}{16} \times \frac{W}{16} \times C}$, where $H$, $W$, $C$ are the height, width, and channel of an input image, this step aims to generate multi-scale local visual features for each query $q \in \mathbb{Q}$. Here, each query $q$ is initialed with a feature embedding $f_q \in \mathbbm{R}^C$, and a position coordinates $(q_x, q_y)$, which are all learnable. Our approach first takes the query $q$ as the center to generate $N_l$ masks $\{M_{q,l}\}_{N_l}$ with the assumption that the scale of one of the masks is $l \times l$. We denote a set $\mathbb{L}$ to contain all the preset scales: $\mathbb{L} = \{l_1,...l,...l_{N_l}\}$. Then, the local visual feature embedding $f_{V,q}^l \in \mathbbm{R}^{(l \times l) \times C}$ is generated as follows:
\begin{equation}
	f_{V,q}^l = \mathcal{S}_V(\mathbf{F}_V, M_{q,l}),
	\label{sampler}
\end{equation}
where $\mathcal{S}_V$ is a feature sampler. For the $l \times l$ elements within $M_{q,l}$, $\mathcal{S}_V$ extracts the corresponding feature embedding from $\mathbf{F}_V$ according to their position coordinates. This generates a local visual embedding $f_{V,q}^l$ to capture the visual representations for $q$ at the scale of $l \times l$. Finally, we integrate all the local embeddings in $f_{V,q}^l$ to calculate an average local feature $\bar{f}_{V,q}^l \in \mathbbm{R}^{C}$ as follows:
\begin{equation}
	\bar{f}_{V,q}^l = \frac{1}{l \times l} \sum_{l \times l} f_{V,q}^l,
	\label{f}
\end{equation}
where $\bar{f}_{V,q}^l$ reflects the average local feature representation at the scale of $l \times l$. For each query q, our approach utilizes $N_l$ masks to generate $N_l$ local feature embedding $\{\bar{f}_{V,q}^l\}_{N_l}$, which well captures multi-scale local feature information around $q$. 

\subsubsection{Depth-guided Scale Matching Prediction} \label{DMS}
As shown in Figure~\ref{SSDA}, given the depth representation $\mathbf{F}_D \in \mathbbm{R}^{\frac{H}{16} \times \frac{W}{16} \times C}$, our approach first employs a depth feature sampler $\mathcal{S}_D$ to directly extract the corresponding feature embedding $f_{D,q} \in \mathbbm{R}^C$ from $\mathbf{F}_D$ for each query $q$:
\begin{equation}
	f_{D,q} = \mathcal{S}_D(\mathbf{F}_D, q).
	\label{sampler2}
\end{equation}
\indent Intuitively, the depth representation $f_{D,q}$ reflects the object scale of a query to some extent, and thus it is reasonable to design a projection function to map $f_{D,q}$ to a scale matching probability distribution $P(q)$, which is formulated as follows:
\begin{equation}
	P(q) = \sigma(f_{D,q}),
	\label{predictor}
\end{equation}
where $\sigma$ is a project function implemented by convolution followed by softmax. $P(q)$ is a $N_l$-dimensional vector to represent the $N_l$ scale matching probabilities, which correspond to the $N_l$ local feature representations $\{\bar{f}_{V,q}^l\}_{N_l}$ in Section~\ref{QMF}.

\begin{figure}[t!]
	\centering
	\includegraphics[scale=0.48]{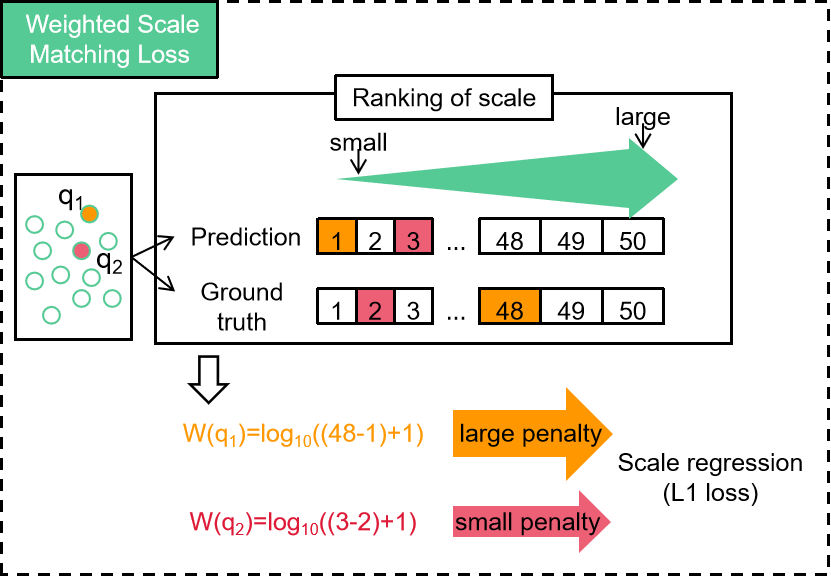}
	\caption{A graphic explanation of WSM loss, where the ranking difference of $q_1$ is large thus the final WSM loss of $q_1$ has been further expanded.}
	\label{wsm}
\end{figure}

After the one-to-one paired Hungarian matching algorithm, all the queries are divided into object queries $q^1$ and non-object queries $q^0$, and we can obtain the ground truth scale $\hat{l}_{q^1}$ (\textit{i.e.}, the width of the 2D box) corresponding to the object queries $q^1$. We expect the predicted scale to be consistent with the true value. Thus, we devise a scale-matching loss to guide the learning of scale prediction:

\begin{equation}
	L(q^1) = \frac{1}{N_l}\sum_{l \in \mathbb{L}}|| P(q^1,l)*l-\hat{l}_{q^1}||_{L_1},
	\label{singleloss}
\end{equation}

where $||.||_{L_1}$ is $L_1$ loss, $P(q^1,l)$ is the predicted probability on the scale $l$, and $P(q^1,l)*l$ represents the predicted scale of $q^1$. $L(q^1)$ reflects the scale prediction error on $q^1$, and different object queries have different error values during training. Generally, queries located on small objects are often prone to generating large errors. To accelerate the model's convergence, it is expected that more attention is paid to the queries with large errors, and thus we assign a penalty weight item $W(q^1)$ to $L(q^1)$ to form our Weighted Scale Matching (WSM) loss, which is expressed as:
\begin{equation}
	L_{WSM} =\frac{1}{\mathbb{B}}\sum_{q^1 \in \mathbb{B}}W(q^1)L(q^1),
	\label{WSMloss}
\end{equation}
where $\mathbb{B}$ represents the number of object queries in one training batch. $W(q^1)$ reflects the importance of object queries, and we adopt a query ranking mechanism to estimate it. Specifically, for all the object queries in $\mathbb{B}$, our approach generates two query ranking queues $Q_P$, $Q_T$ in descending order according to their predicted scales and true scales, respectively, and $W(q^1)$ is calculated as:
\begin{equation}
	W(q^1) =log(|Index(Q_T,q^1)-Index(Q_P,q^1)| + 1),
	\label{Weight}
\end{equation}
where $Index(Q,q^1)$ represents the ranking index of $q^1$ in $Q$. When $q^1$ has the same index in both queues, $W(q^1)=0$. Otherwise, the large index difference between $Index(Q_T,q^1)$ and $Index(Q_P,q^1)$ indicates that the predicted scale has a large error, and thus it is required to give a high penalty weight on $q^1$. A graphic explanation of WSM loss is shown in Figure \ref{wsm}. Compared to the previous loss weighting approaches \cite{huang2020epnet, ronneberger2015u}, our ranking mechanism considers the relative weighting correlations among all the object queries in a training batch, which is more reasonable and global-aware. \\
\indent As it can be observed, all the masks are preset to squares and we use the width labels of 2D boxes to supervise the scale of masks. As shown in Figure~\ref{intro}(a), suffered by the similar features of the surrounding objects, the key points of a query are very easily located on other objects without any supervision in the deformable attention layer. From the perspective of verity, all the objects are on the same horizontal plane, thus in a 2D image, the nearby objects can only appear on the left or right of the target object, but not on the top or bottom. To this end, only limiting and designing a loss to learn the width of the preset masks and restricting the height to be the same, can help reach a desirable balance between the performance and learning cost of the network.

\subsubsection{Object Query Updating} \label{update}
The same as Deformable DETR~\cite{zhu2020deformable}, given an initial query, the query updating aims to search for several key points around the query, and then integrate these key points with attention weights to form the new query feature for the following attributes detection. As aforementioned, although MonoDETR utilizes both depth and visual features to predict key points, there still exist many noise key points.
To alleviate this problem, as shown in Figure~\ref{SSDA}, our query updating utilizes both multi-scale local visual features and scale-matching probability to construct a scale-aware filter to help find more accurate key points. Concretely, given the multi-scale local feature and the scale-matching probability of a query, our approach first multiplies them and then employs an adaptive learning layer to generate a scale-aware filter. Here, the adaptive learning layer contains a lightweight convolutional network with a batch normalization layer and an active layer. The generation process of scale-aware filter $\mathcal{F} \in \mathbbm{R}^{N \times C}$ can be expressed as:
\begin{equation}
	\mathcal{F} = Relu(BN(Conv(\sum_{l=1}^L\sum_{q \in \mathbb{Q}}{\bar{f}_{V,q}^l * P(q, l)}))).
	\label{filter}
\end{equation}

Next, we execute the element-wise operation between the filter and the initial query feature to generate the scale-aware features to better predict key points around their relative queries. For better convergence, we do not directly predict the coordinate of the key point but predict the offset $\triangle p_q = (\triangle q_x, \triangle q_y)$ from the query coordinate $p_q = (q_x, q_y)$:
\begin{equation}
	\triangle p_q = Linear(\mathcal{F} \cdot f_q),
\label{kp}
\end{equation}
where the $Linear$ operation project the input $C$-dimensional feature into $K \times 2$-dimensional coordinate vector, and $K$ is the number of key point for each query.
Finally, the query feature is updated by integrating these key points features with the attention weights. The updated feature is obtained by: 
\begin{equation}
\begin{aligned}
\label{equ_update}
&FeaUpd(\mathcal{F}, p_q) \\
 &= \sum_{m=1}^M W_m[\sum_{k=1}^K A_{mqk} \cdot W_m^{’} \mathbf{F}_V(p_q + \triangle p_{mqk}) ],
\end{aligned}
\end{equation}
where $M$ is the head number of attention layer, $W_m, W_m^{’}$ are the learnable weights, $A_{mqk}$ is the attention weights. $\mathbf{F}_V(p_q + \triangle p_{mqk})$ presents the key point features sampled on visual feature map $\mathbf{F}_V$ by the bilinear interpolation.

%Benefiting from the scale-aware filter, our approach could predict accurate key points to better support query feature generation for 3D attribute prediction.

\subsection{Training Loss} \label{loss}
SSD-MonoDETR is an end-to-end network, and all the components are jointly trained according to a composite loss function, which consists of $L_{2D}$, $L_{3D}$ and  $L_{WSM}$. Specifically, the 2D object loss $L_{2D}$ uses Focal loss \cite{lin2017focal} to estimate the object classes, L1 loss to estimate the 2D sizes $(l, r, t, b)$ and projected 3D center $(x_{3d}, y_{3d})$, and GIoU loss~\cite{rezatofighi2019generalized} for 2D box IoU. Finally, the $L_{2D}$ could be expressed as:
\begin{equation}
	L_{2D} =  \lambda_1 L_{class} + \lambda_2 L_{2dsize} + \lambda_3 L_{xy3d} + \lambda_4 L_{giou}.
	\label{2dloss}
\end{equation}
\indent As for the 3D object loss $L_{3D}$, we follow MonoDLE~\cite{ma2021delving} to predict 3D sizes $(h_{3d}, w_{3d}, l_{3d})$ and orientation angle $\alpha$. For the depth prediction, we use the average of three depth values to present the predicted depth $d_{pre}$:
\begin{equation}
	d_{pre} =  \frac{1}{3}(d_{reg} + d_{geo} + d_{map}),
	\label{dpre_first}
\end{equation}
where $d_{reg}$ is the depth value directly regressed by one detection head, $d_{geo}$ is obtained by the  relationship between 2D and 3D sizes:
\begin{equation}
	d_{geo} =  f\frac{h_{3D}}{t+b},
	\label{dpre_second}
\end{equation}
where $f$ is the camera focal length. $d_{map}$ get value by the projected 3D center $(x_{3d}, y_{3d})$ and interpolation algorithm on the depth map (see MonoDETR~\cite{zhang2022monodetr} for more details). For the whole $d_{pre}$, a Laplacian aleatoric uncertainty loss \cite{chen2020monopair} is adopted to form the final depth loss:
\begin{equation}
	L_{depth} =  \frac{2}{\sigma}||d_{gt} - d_{pre}||_1 + log(\sigma),
	\label{loss_d}
\end{equation}
where $\sigma$ is the the standard deviation predicted together with $d_{reg}$, $d_{gt}$ is the ground truth depth value. As a whole, the $L_{3D}$ could be expressed as:
\begin{equation}
	L_{3D} =  \lambda_5 L_{3dsize} + \lambda_6 L_{angle} + \lambda_7 L_{depth}.
	\label{3dloss}
\end{equation}
\indent Additional, the proposed WSM loss $L_{WSM}$ is equipped on the SSDA module to improve query features. As a result, our composite loss is expressed as:
\begin{equation}
	L = L_{2D} + L_{3D} + \lambda_8 L_{WSM},
	\label{all loss}
\end{equation}
where $\lambda_1$ to $\lambda_8$ are the balancing weights. For $\lambda_8$, we will conduct its utility evaluation in the experiments.

\begin{table*}[t!]
	\begin{center}
		\begingroup
		\setlength{\tabcolsep}{0.5pt} % Default value: 6pt
		\renewcommand{\arraystretch}{1.2} % Default value: 1
  \caption{Performance of the ``Car'' category on KITTI test and val sets, where ``v.$\checkmark$'' and ``v.\ding{55}'' represent the accuracy improvement compared to the methods with/without extra data. The first/second best results are highlighted in \textcolor{red}{red} / \textcolor{blue}{blue}  fonts and the improvements are noted in bold fonts.}
  \label{test}
		\begin{tabular}{|m{2.7cm}<{\centering}|m{1.9cm}<{\centering}|m{0.95cm}<{\centering}m{0.95cm}<{\centering}m{0.95cm}<{\centering}|m{0.95cm}<{\centering}m{0.95cm}<{\centering}m{0.95cm}<{\centering}|m{0.95cm}<{\centering}m{0.95cm}<{\centering}m{0.95cm}<{\centering}|m{0.95cm}<{\centering}m{0.95cm}<{\centering}m{0.95cm}<{\centering}|m{0.8cm}<{\centering}|}
			\hline
			\multirow{2}{*}{Method} &
			\multirow{2}{*}{Reference} &
			\multicolumn{3}{c|}{Test, $AP_{3D}$ }&
			\multicolumn{3}{c|}{Test, $AP_{BEV}$ }&
			\multicolumn{3}{c|}{Val, $AP_{3D}$ }&
			\multicolumn{3}{c|}{Val, $AP_{BEV}$ }&
			\multirow{2}{*}{\makecell[c]{Time\\(ms)}} \\
			{ } & { }&Easy & Mod. & Hard & Easy & Mod. & Hard &Easy & Mod. & Hard &Easy & Mod. & Hard & {}\\
			\hline\hline
			With extra data:&&&&&&&&&&&&&&\\
			MonoRUn \cite{chen2021monorun} & CVPR2021&19.65 &12.30 &10.58 &27.94 &17.34 &15.24 &20.02&14.65 &12.61 &- & - & - &70\\
			DDMP-3D \cite{wang2021depth} &CVPR2021& 19.71 &12.78 &9.80 &28.08 &17.89 &13.44& -& -& -&- & - & - &180\\
			CaDDN \cite{reading2021categorical} & CVPR2021&19.17 &13.41 &11.46 &27.94 &18.91 &17.19 &23.57 &16.31 &13.84 &- & - & - &630\\
			AutoShape \cite{liu2021autoshape} &ICCV2021& 22.47 &14.17 &11.36& 30.66& 20.08 &15.59 &20.09& 14.65& 12.07 &- & - & - &-\\
			MonoDTR \cite{huang2022monodtr}  &CVPR2022& 21.99 &15.39 &12.73 &28.59 &20.38 &17.14 &24.52 &18.57 &15.51&33.33 &25.35&21.68&37\\
			DID-M3D \cite{peng2022did} & ECCV2022 &24.40 &16.29 &13.75& 32.95&22.76& 19.83 &22.98& 16.12& 14.03 &31.10 &22.76 &19.50&40\\
			OPA-3D \cite{su2023opa} & RAL2023 &24.60 &17.05&\textcolor{blue}{14.25} &\textcolor{blue}{33.54} &22.53 &19.22&24.97 &19.40&\textcolor{blue}{16.59}&33.80 &25.51&22.13 &40\\
   MonoPGC \cite{wu2023monopgc} &ICRA2023&\textcolor{blue}{24.68} &\textcolor{blue}{17.17} &14.14 &32.50 &\textcolor{blue}{23.14} &\textcolor{blue}{20.30}&25.67 &18.63 &15.65 &34.06 &24.26 &20.78&46\\
			\hline
			Without extra data:&&&&&&&&&&&&&&\\
			MonoGeo \cite{zhang2021learning} & CVPR2021 &18.85 &13.81 &11.52 &25.86 &18.99 &16.19& 18.45 &14.48 &12.87&27.15 &21.17 &18.35&50\\
			MonoFlex \cite{zhang2021objects} & CVPR2021&19.94 &13.89 &12.07 &28.23 &19.75 &16.89 &23.64 &17.51 &14.83 &- & - & - &30\\
			GUPNet \cite{lu2021geometry} & ICCV2021 &20.11 &14.20 &11.77& -& -& - &22.76& 16.46& 13.72 &31.07 &22.94 &19.75&-\\
			DEVIANT \cite{kumar2022deviant} &ECCV2022 &21.88 &14.46 &11.89 &29.65 &20.44 &17.43 &24.63 &16.54 &14.52 &32.60 &23.04 &19.99&40\\
   MonoDETR \cite{zhang2022monodetr} & CVPR2022&23.65 &15.92 &12.99 &32.08 &21.44 &17.85 &\textcolor{red}{28.84} &\textcolor{blue}{20.61}&16.38&\textcolor{blue}{37.86}& \textcolor{blue}{26.95}&\textcolor{blue}{22.80}&\textcolor{red}{20}\\
               MonoJSG \cite{lian2022monojsg} & CVPR2022&\textcolor{red}{24.69}& 16.14 &13.64 &32.59 &21.26& 18.18 &26.40 &18.30 &15.40&- & - & - &42\\
   MonoRCNN++~\cite{shi2023multivariate} &WACV2023& 20.08 &13.72& 11.34&-&-&-&19.07&14.87&12.59&26.41&20.80&17.27&-\\
MonoEdge~\cite{zhu2023monoedge} &WACV2023&21.08 &14.47 &12.73 &28.80 &20.35 &17.57 &25.66 &18.89 &16.10 &33.71 &25.35 &22.18 &37\\
PDR~\cite{sheng2023pdr} &TCSVT2023&23.69 &16.14 &13.78 &31.76 &21.74 &18.79 &27.65 &19.44 &16.24 &35.59 &25.72 &21.35&29\\
			SSD-MonoDETR & - &24.52&\textcolor{red}{17.88} &\textcolor{red}{15.69} &\textcolor{blue}{33.59} &\textcolor{red}{24.35} &\textcolor{red}{21.98} &\textcolor{blue}{29.53} &\textcolor{red}{21.96} &\textcolor{red}{18.20} &\textcolor{red}{38.00}&\textcolor{red}{29.44}& \textcolor{red}{26.94}&\textcolor{blue}{21}\\
			\hline
			Improvement & \textit{vs.} Extra \checkmark &-&$\bm{0.71}$&$\bm{1.44}$&$\bm{0.05}$&$\bm{1.21}$& $\bm{1.68}$& $\bm{4.56}$&$\bm{2.56}$&$\bm{1.61}$&$\bm{4.20}$ & $\bm{3.93}$ & $\bm{4.81}$ &-\\
			Improvement & \textit{vs.} Extra $\times$ &-&$\bm{1.41}$&$\bm{2.11}$&-&$\bm{2.24}$& $\bm{3.38}$& $\bm{0.69}$&$\bm{1.35}$&$\bm{1.82}$&$\bm{0.14}$ & $\bm{2.49}$&$\bm{4.14}$ &-\\
			\hline
		\end{tabular}
		\endgroup
	\end{center}
\end{table*}

\begin{table}[t!]
	\begin{center}
		\begingroup
		\setlength{\tabcolsep}{0.5pt} % Default value: 6pt
		\renewcommand{\arraystretch}{1.2} % Default value: 1
		%\begin{tabular}{c|c|ccc|ccc|ccc|ccc}
  \caption{Performance of the ``Pedestrian'' and ``Cyclist'' categories on KITTI test set.}
  \label{test2}
		\begin{tabular}{|m{2.7cm}<{\centering}|m{0.95cm}<{\centering}m{0.95cm}<{\centering}m{0.95cm}<{\centering}|m{0.95cm}<{\centering}m{0.95cm}<{\centering}m{0.95cm}<{\centering}|}
			\hline
			\multirow{2}{*}{Method} &
			\multicolumn{3}{c|}{Pedestrian, $AP_{3D}$}&
			\multicolumn{3}{c|}{Cyclist, $AP_{3D}$} \\
			{ } & Easy & Mod. & Hard & Easy & Mod. & Hard \\
			\hline\hline
			With extra data:&&&&&&\\
            D4LCN \cite{ding2020learning}  &4.55 &3.42 &2.83 &2.45 &1.67 &1.36 \\
            MonoEF \cite{zhou2021monoef} & 4.27 &2.79 &2.21 &1.80 &0.92 &0.71\\
            DDMP-3D \cite{wang2021depth} &4.93 &3.55 &3.01 &4.18 &2.50& 2.32\\
            DFR-Net \cite{zou2021devil} & 6.09 &3.62 &3.39 &5.69 &3.58 &3.10 \\
		MonoPSR \cite{ku2019monocular} & 6.12 &4.00 &3.30& \textcolor{red}{8.70} &\textcolor{blue}{4.74}& \textcolor{blue}{3.68} \\
            CADNN \cite{reading2021categorical} &12.87 &8.14 &6.76 &7.00& 3.41 &3.30\\
            MonoPGC \cite{wu2023monopgc} &\textcolor{red}{14.16} &\textcolor{blue}{9.67} &\textcolor{blue}{8.26} &5.88 &3.30 &2.85\\
			\hline
			Without extra data:&&&&&&\\
            MonoGeo \cite{zhang2021learning} &8.00 &5.63 &4.71 &4.73 &2.93 &2.58\\
			MonoFlex \cite{zhang2021objects} & 9.43 &6.31 &5.26& 4.17 &2.35& 2.04 \\
            MonoDLE \cite{ma2021delving} & 9.64 &6.55 &5.44 &4.59 &2.66 &2.45 \\
            MonoPair \cite{chen2020monopair} &10.02 &6.68 &5.53& 3.79 &2.12 &1.83\\
            PDR~\cite{sheng2023pdr} &11.61 &7.72 &6.40 &2.72 &1.57 &1.50\\
			MonoDETR \cite{zhang2022monodetr}&12.54& 7.89& 6.65& 7.33& 4.18& 2.92 \\
   MonoRCNN++~\cite{shi2023multivariate} &12.26&7.90&6.62 &3.17&1.81&1.75\\
			DEVIANT \cite{kumar2022deviant} & \textcolor{blue}{13.43} &8.65 &7.69 &5.05 &3.13 &2.59\\
   SSD-MonoDETR &12.64&$\textcolor{red}{9.88}$ &$\textcolor{red}{8.58}$ &$\textcolor{blue}{7.79}$ &$\textcolor{red}{5.76}$ & $\textcolor{red}{4.33}$ \\
			\hline
			Improvement v.\checkmark &-& $\bm{0.21}$&$\bm{0.32}$&-& $\bm{1.02}$ & $\bm{0.65}$ \\
			Improvement v.\ding{55} &-& $\bm{1.23}$ & $\bm{0.89}$ &$\bm{0.46}$& $\bm{1.58}$ & $\bm{1.41}$ \\
			\hline
		\end{tabular}
		\endgroup
	\end{center}
\end{table}

\begin{table*}[t!]
	\begin{center}
    \begingroup
    \setlength{\tabcolsep}{0.5pt} 
    \renewcommand{\arraystretch}{1.2} 
	\caption{Vehicle performance on the Waymo Open Val set between different models. We use $AP_{3D}$ ($LEVEL\_1$ and $LEVEL\_2$, IoU \textgreater 0.5 and IoU \textgreater 0.7) according to three object distance intervals.}
 \label{waymo}
\begin{tabular}{|m{2.5cm}<{\centering}|m{1.3cm}<{\centering}|m{0.8cm}<{\centering}m{0.8cm}<{\centering} m{0.9cm}<{\centering}m{0.94cm}<{\centering}|m{0.8cm}<{\centering}m{0.8cm}<{\centering} m{0.9cm}<{\centering}m{0.94cm}<{\centering}|m{0.8cm}<{\centering}m{0.8cm}<{\centering} m{0.9cm}<{\centering}m{0.94cm}<{\centering}|m{0.8cm}<{\centering}m{0.8cm}<{\centering} m{0.9cm}<{\centering}m{0.94cm}<{\centering}|}
\hline
\multirow{2}{*}{Method} &
\multirow{2}{*}{Reference} &
\multicolumn{4}{c|}{$LEVEL\_1$(IoU \textgreater 0.5)}&
\multicolumn{4}{c|}{$LEVEL\_2$(IoU \textgreater 0.5)}&
\multicolumn{4}{c|}{$LEVEL\_1$(IoU \textgreater 0.7)}&
\multicolumn{4}{c|}{$LEVEL\_2$(IoU \textgreater 0.7)}\\
{ } &{ }&Overall &0-30m &30-50m & 50m-Inf&Overall &0-30m &30-50m & 50m-Inf&Overall &0-30m &30-50m & 50m-Inf&Overall &0-30m &30-50m & 50m-Inf \\
\hline\hline
With extra data:&&&&& &&&& &&&& &&&&\\
PatchNet~\cite{ma2020rethinking} &ECCV 20 &2.92 &10.03 &1.09 &0.23 &2.42 &10.01 &1.07 &0.22& 0.39 &1.67 &0.13 &0.03 &0.38 &1.67 &0.13 &0.03\\
PCT~\cite{wang2021progressive} &NIPS 21 & 4.20 &14.70 &1.78 &0.39 &4.03 &14.67 &1.74 &0.36& 0.89 &3.18 &0.27 &0.07&0.66 &3.18 &0.27 &0.07 \\
\hline
Without extra data:&&&&& &&&& &&&& &&&&\\
GUPNet~\cite{lu2021geometry} &ICCV 21& 10.02 &24.78 &4.84 &0.22 &9.39 &24.69 &4.67 &0.19& 2.28 &6.15 &0.81 &0.03& 2.14 &6.13 &0.78& 0.02\\
MonoJSG~\cite{lian2022monojsg} &CVPR 22& 5.65 &20.86 &3.91 &$\textcolor{red}{0.97}$ &5.34 &20.79 &3.79 &$\textcolor{red}{0.85}$&0.97 &4.65 &0.55 &$\textcolor{blue}{0.10}$ &0.91 &4.64 &0.55 &$\textcolor{blue}{0.09}$\\
DEVIANT~\cite{kumar2022deviant} &ECCV 22 &10.98 &26.85 &$\textcolor{blue}{5.13}$ &0.18 &10.29 &26.75 &$\textcolor{blue}{4.95}$ &0.16&2.69 &6.95 &$\textcolor{blue}{0.99}$ &0.02 &2.52 &6.93 &$\textcolor{blue}{0.95}$ &0.02\\
MonoRCNN++~\cite{shi2023multivariate} &WACV 23&$\textcolor{blue}{11.37}$ &$\textcolor{red}{27.95}$ &4.07 &0.42 &$\textcolor{blue}{10.79}$ &$\textcolor{red}{27.88}$ &3.98& 0.39&$\textcolor{blue}{4.28}$ &$\textcolor{blue}{9.84}$ &0.91 &0.09 &$\textcolor{blue}{4.05}$ &$\textcolor{red}{9.81}$ &0.89 &0.08\\
SSD-MonoDETR& - &$\textcolor{red}{11.83}$&$\textcolor{blue}{27.69}$ &$\textcolor{red}{5.33}$ &$\textcolor{blue}{0.85}$& $\textcolor{red}{11.34}$&$\textcolor{blue}{27.62}$ &$\textcolor{red}{5.21}$&$\textcolor{blue}{0.76}$&$\textcolor{red}{4.54}$ &$\textcolor{red}{9.93}$&$\textcolor{red}{1.18}$ &$\textcolor{red}{0.15}$&$\textcolor{red}{4.12}$ &$\textcolor{blue}{8.87}$&$\textcolor{red}{1.02}$&$\textcolor{red}{0.13}$ \\
\hline
\end{tabular}
\endgroup
\end{center}
\end{table*}

\section{Experiments} \label{exp}
\subsection{Experimental Setup}
\textbf{Dataset:} We verify the proposed SSD-MonoDETR on the popular KITTI dataset~\cite{geiger2012we} and Waymo Open dataset~\cite{ettinger2021large}. KITTI includes $7,481$ training and $7,518$ testing images. Considering the unseen labels on the testing set, we follow~\cite{chen20153d} to further split the training samples into $3,712$ samples for the sub-training set and $3,769$ samples for the validation set. We measure the detection results on three-level difficult samples (easy, moderate, and hard), and mainly evaluate the performance on the class ``Car'' by using the average precision (AP) in 3D space and the bird-eye view denoted as AP$_{3D}$ and AP$_{BEV}$, respectively, which are at $40$ recall positions. Also, the results of the class ``Pedestrian'' and ``Cyclist'' are listed by using the average precision (AP) in 3D space to make a comparison with the other existing approaches. \\
\indent Waymo Open is a larger and more challenging autonomous driving scene understanding dataset, which contains $798$ training and $202$ validation sequences and generates nearly $160k$ and $40k$ samples, respectively. For training and testing our method, we follow~\cite{kumar2022deviant} to generate $52,386$ training and $39,848$ validation images from the front camera, and the images for training are constructed by sampling every third frame from the training sequences. For the evaluation metric, the Waymo Open dataset defines two difficulty levels ($Level\_1$: points on object $\geq$ 5, and $Level\_2$: points on object $\geq$ 1) for all the instances, and each difficulty adopts two thresholds ($0.5, 0.7$) and four distance ranges ($Overall, 0-30m, 30-50m, 50m-inf$) to evaluate the detection results.

\textbf{Implementation Details:}
We adopt ResNet-50~\cite{he2016deep} as the feature extraction backbone and all the attention layers have $8$ heads.
For the decoder, we set $3$ blocks and the number $N$ of input queries is set to $50$. We select $5$ scales $\{1, 3, 5, 7, 9\}$ of masks to extract local features for ``Car''.
For ``Pedestrian'' and ``Cyclist'', whose aspect ratio (h/w) are larger than ``Car'' but have a relatively small size, to extract more accurate local features, we set the scales to $\{1, 3, 5\}$ and expand the vertical coordinates of mask elements to $3$ and $2$ times respectively. For the balancing weights $\lambda_1$ to $\lambda_7$ in Section~\ref{loss}, we follow the settings in MonoDETR, which are: $\{2, 10, 5, 2, 1, 1, 1\}$. We set the $\lambda_8$ to $0.2$ according to the following evaluation experiment. Our training is conducted on a single RTX A6000 GPU by using the Adam optimizer with a weight decay $10^{-4}$ for $200$ epochs, where the batch size is $16$ and the learning rate is $2 \times 10^{-4}$. Please note that all experimental results involving ``Car'', ``Pedestrian'', and ``Cyclist'' are obtained using a single-category training approach. For Waymo Open, we list the results of the ``Vehicle'' (\textit{i.e.}, ``Car'')  for comparison with other state-of-the-art methods, and the setting of masks is the same as KITTI. We also adopt the Adam optimizer with a weight decay of $10^{-4}$ to train our model for $40$ epochs, the batch size of $40$, and the learning rate of $5 \times 10^{-4}$.

\subsection{Performance Comparison}

Table~\ref{test} shows the comparison results of the ``Car'' category on the KITTI test and validation sets.
We compare our approach with several advanced approaches by using or not using extra data. From the results, we reach the following conclusions:
\begin{enumerate}
	\item Among all the approaches, SSD-MonoDETR achieves the best accuracies on moderate and hard objects, with $0.83\% \sim 4.81\%$ improvements. This improvement accelerates as sample difficulty increases, which is consistent with the characteristics of our method. This is because hard samples usually have small sizes, and thus the estimated query points by the previous algorithms are prone to deviate from objects, resulting in detection errors. Comparatively, SSDA well estimates the scales of query points by using depth-guided scale matching prediction, bringing a significant performance improvement.
	\item  We discover that our approach achieves greater performance improvements on AP$_{BEV}$ as compared to AP$_{3D}$. This is because AP$_{BEV}$ mainly focuses on evaluating the relative positions of cars with respect to streets in bird view, which is highly related to the positions of the estimated query points. Comparatively, AP$_{3D}$ pays more attention to size parameters inside objects like the distance assessment from object center to ground, which is less relative to the query points.
	\item Our approach has a similar testing speed as compared to MonoDETR \cite{zhang2022monodetr}, but achieves better accuracy, especially on moderate and hard objects. Although SSDA brings extra computation costs, our approach reduces a cross-attention operation as compared to MonoDETR, resulting in a similar testing speed.
    \item On the ``Easy'' subset, our method does not perform as well as the `Moderate'' and ``Hard'' subsets. This stems from the difficulty classification criteria of the KITTI dataset, where the ``Easy'' objects are mostly close and their degree of occlusion by other objects is $0$. Thus, the ``Easy'' objects usually have high-quality image features and are slightly suffered by the surrounding objects, which results in the proposed SSDA not being able to present an obvious effect as it does in the ``Moderate'' and ``Hard'' subsets.
\end{enumerate}

\indent Table~\ref{test2} further shows the performance on ``Pedestrian'' and ``Cyclist'' categories.
It is evident that these two categories present greater challenges compared to the ``Car'' category, mainly due to their smaller size and non-rigid body nature. 
Thanks to the effective SSDA module, our method surpasses all the previous methods without extra data in terms of performance across three difficulty levels, particularly for moderate and hard objects, where we achieve an approximate $2\%$ improvement on ``Pedestrian'' and $1.5\%$ improvement on ``Cyclist''.
The results presented in Table~\ref{test2} validate the exceptional generality and scalability of our model, which only relies on easily accessible prior knowledge about different categories of scales. As a result, our proposed method effortlessly achieves accurate detection of objects with diverse appearances. \\
\indent Table~\ref{waymo} lists the average precision on 3D view ($AP_{3D}$) results on ``Vehicle'' of different methods on the Waymo Open Val set. On all the difficulty levels and IoU thresholds, our SSD-MonoDETR yields superior breakthroughs against all the other monocular 3D detectors in terms of ``Overall'' perspective, which proves the effectiveness of our proposed SSDA layer and WSM loss. On the distance $30m-inf$ with the strict IoU threshold of $0.7$, our method also exceeds all the other methods, which is consistent with the design motivation of our proposed scale-aware mechanism, generating higher-quality query features thus obviously improving the accuracy for those hard and distant objects.

\begin{table}[t!]
	\begin{center}
		\renewcommand{\arraystretch}{1.2}
            \caption{Performance change by using different scale settings.}
		\setlength{\tabcolsep}{2.5pt}{
			\begin{tabular}{|m{3cm}<{\centering}|m{1.5cm}<{\centering}m{1.5cm}<{\centering}m{1.5cm}<{\centering}|}
				\hline
				Multi-scale & Easy & Mod. & Hard \\
				\hline\hline
				$\{3,5,7\}$ & 28.01 & 20.16 & 17.03 \\
				$\{3,5,7,9\}$ & 29.22 & 21.67 & 17.40 \\
				$\{1,3,5,7,9\}$ & $\bm{29.53}$ & $\bm{21.96}$ & $\bm{18.20}$  \\
				$\{1,3,5,7,9,11\}$ & 29.26& 20.88 & 17.55\\
				$\{1,3,5,7,9,11,13\}$ & 27.50& 19.79& 16.02\\
				\hline
		\end{tabular}}
		\label{ab2}
	\end{center}
\end{table}

\subsection{Evaluation on SSDA}
In this experiment, we first try different scale settings in SSDA to observe the performance trend and then evaluate the quality of the generated query points by SSDA.

\textbf{Multi-scale settings in SSDA}: Before SSDA, the input images are resized to $1/16$ during feature extraction, and thus most objects are reduced within $1$ to $10$ pixels in scale. Therefore, we set the scale value in the range of $[1,13]$ to observe the performance trend, which is shown in Table~\ref{ab2}. Initially, we set three masks in SSDA with the scales $\{3, 5, 7\}$, which achieves unsatisfactory performance because many object scales are not covered. Then, we add one more mask with a scale of $9$ and discover that the performances on easy and moderate objects are significantly improved.
On this basis, the further introduction of the scale $1$ boosts the performance, especially on hard objects, which usually have small sizes.
Moreover, the performance begins to degrade as the scales $11$ and $13$ are added.
We attribute this to that these scales exceed the sizes of most objects and would interrupt the scale estimation on object queries.\\

\begin{table}[t!]
	\begin{center}
		\renewcommand{\arraystretch}{1.2}
  \caption{The prediction precision of key points by MonoDETR, our method without WSM loss, and our complete method.}
		\setlength{\tabcolsep}{2.5pt}{
			\begin{tabular}{|m{3.5cm}<{\centering}|m{1.5cm}<{\centering}m{3cm}<{\centering}|}
				\hline
				Method              &  Position Precision & Weighted Position Precision \\
				\hline\hline
				MonoDETR            & 70.17 $\%$  & 74.61$\%$  \\
				SSD-MonoDETR/WSM   & 73.56 $\%$ & 86.13 $\%$  \\
				SSD-MonoDETR                  & $\bm{76.34\%}$ & $\bm{94.78\%}$  \\
				\hline
		\end{tabular}}
		\label{ER}
	\end{center}
\end{table}

\begin{figure*}[t!]
	\centering
	\includegraphics[scale=0.33]{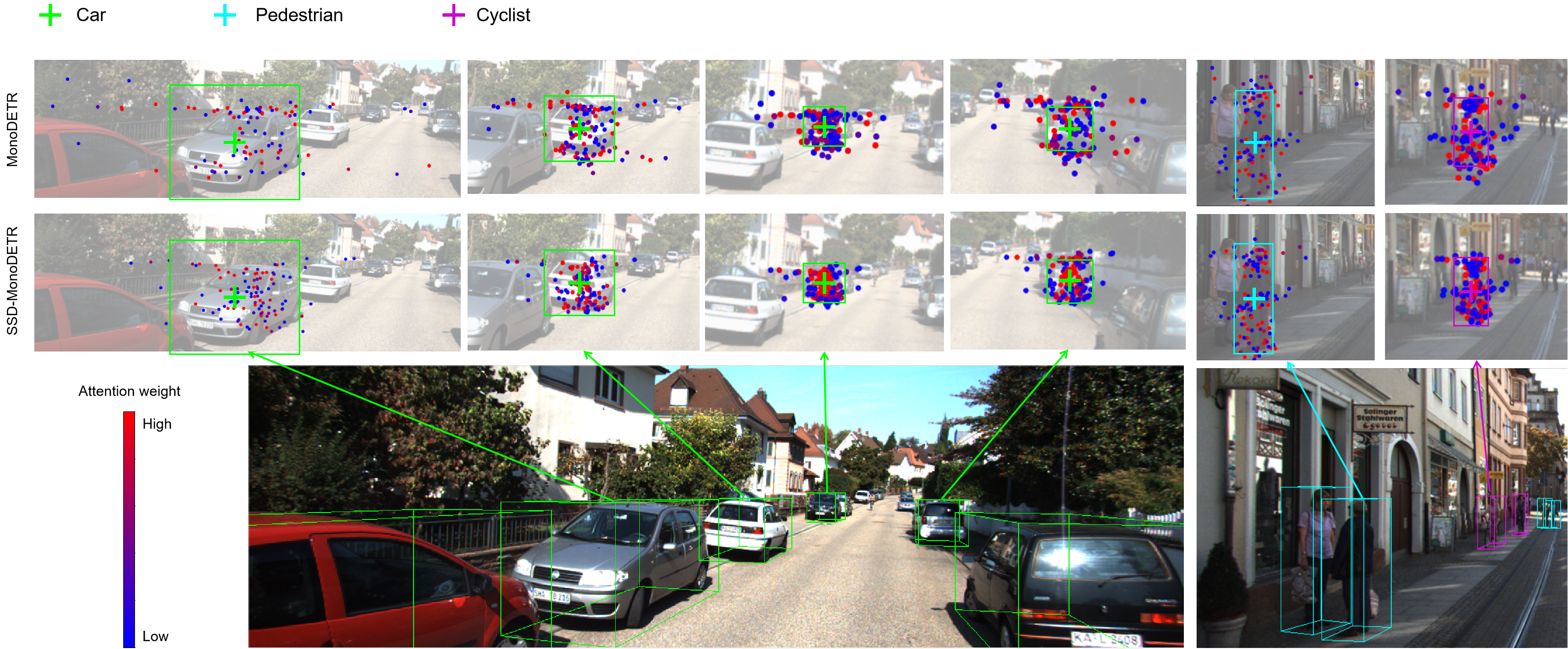}
	\caption{An example to illustrate the distribution of the generated key points by MonoDETR and our SSD-MonoDETR.}
	\label{point}
\end{figure*}

\begin{figure*}[ht]
	\centering
	\begin{minipage}{0.495\linewidth}
		\centering
		\includegraphics[width=0.98\linewidth]{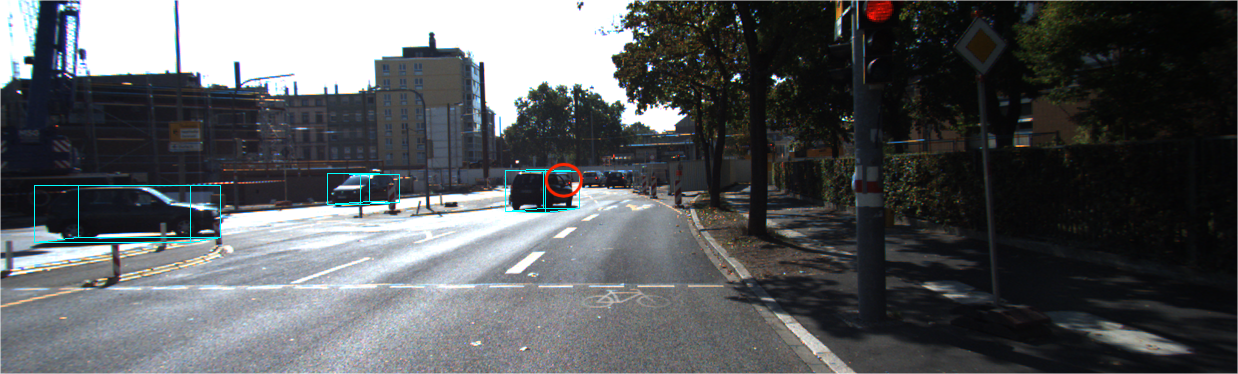}\vspace{8pt}
	\end{minipage}
	\begin{minipage}{0.495\linewidth}
		\centering
		\includegraphics[width=0.98\linewidth]{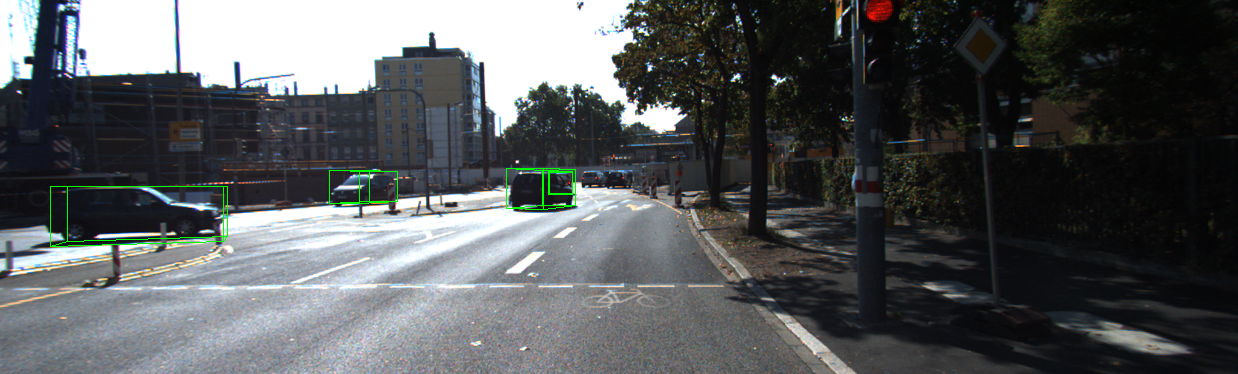}\vspace{8pt}
	\end{minipage}
	\begin{minipage}{0.495\linewidth}
		\centering
		\includegraphics[width=0.98\linewidth]{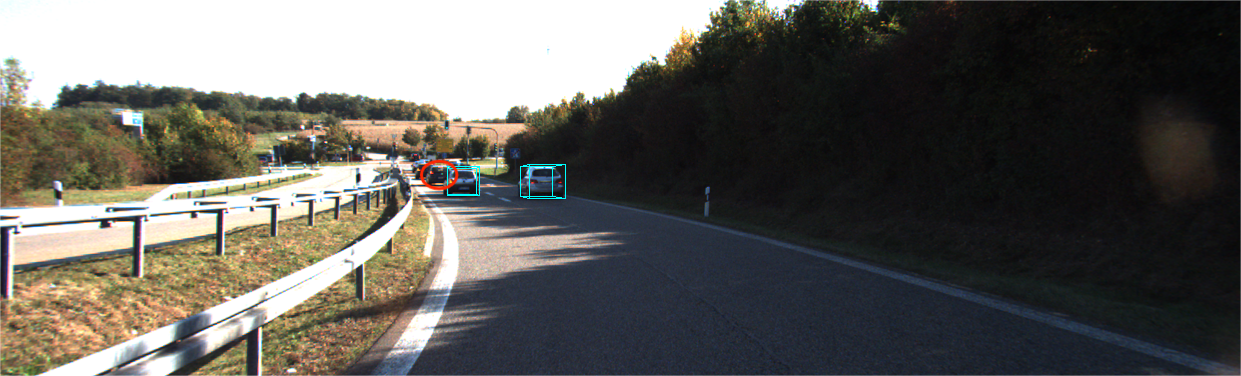}\vspace{8pt}
	\end{minipage}
	\begin{minipage}{0.495\linewidth}
		\centering
		\includegraphics[width=0.98\linewidth]{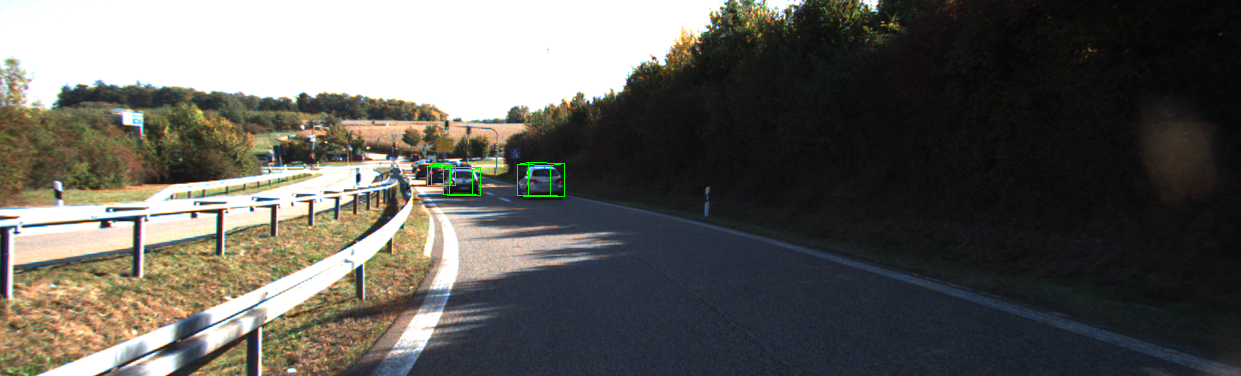}\vspace{8pt}
	\end{minipage}
	\begin{minipage}{0.495\linewidth}
		\centering
		\includegraphics[width=0.98\linewidth]{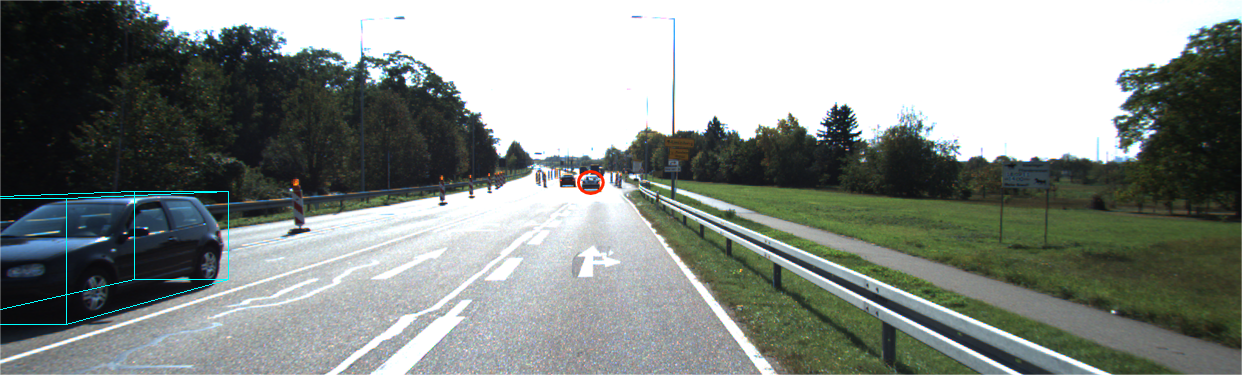}\vspace{8pt}
	\end{minipage}
	\begin{minipage}{0.495\linewidth}
		\centering
		\includegraphics[width=0.98\linewidth]{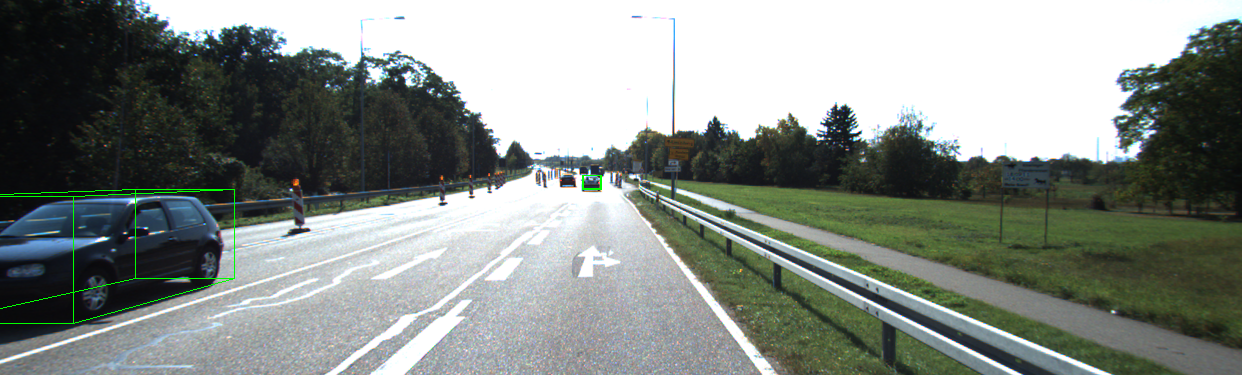}\vspace{8pt}
	\end{minipage}
	\begin{minipage}{0.495\linewidth}
		\centering
		\includegraphics[width=0.98\linewidth]{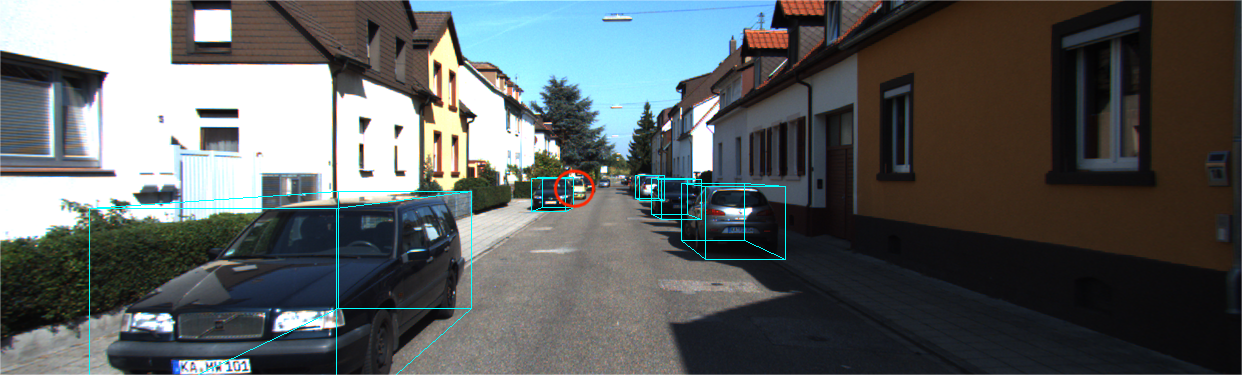}\vspace{8pt}
	\end{minipage}
	\begin{minipage}{0.495\linewidth}
		\centering
		\includegraphics[width=0.98\linewidth]{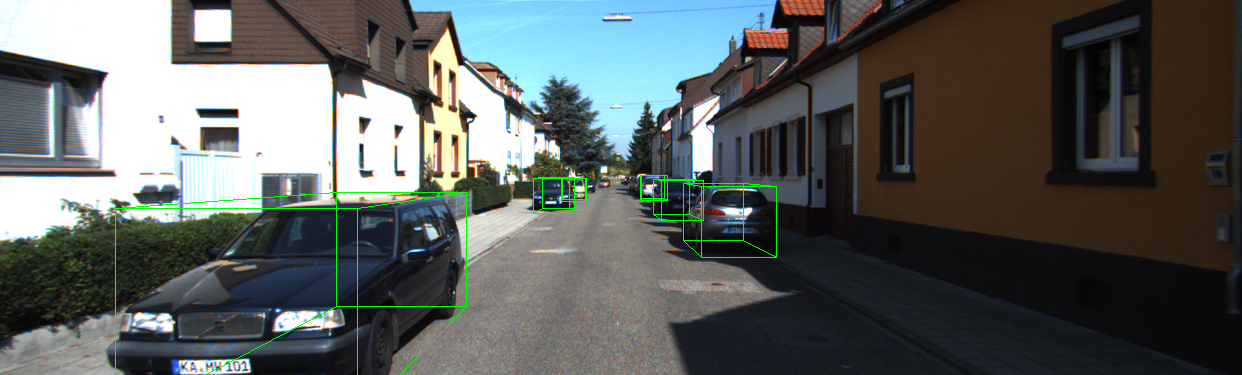}\vspace{8pt}
	\end{minipage}
    \begin{minipage}{0.495\linewidth}
		\centering
		\includegraphics[width=0.98\linewidth]{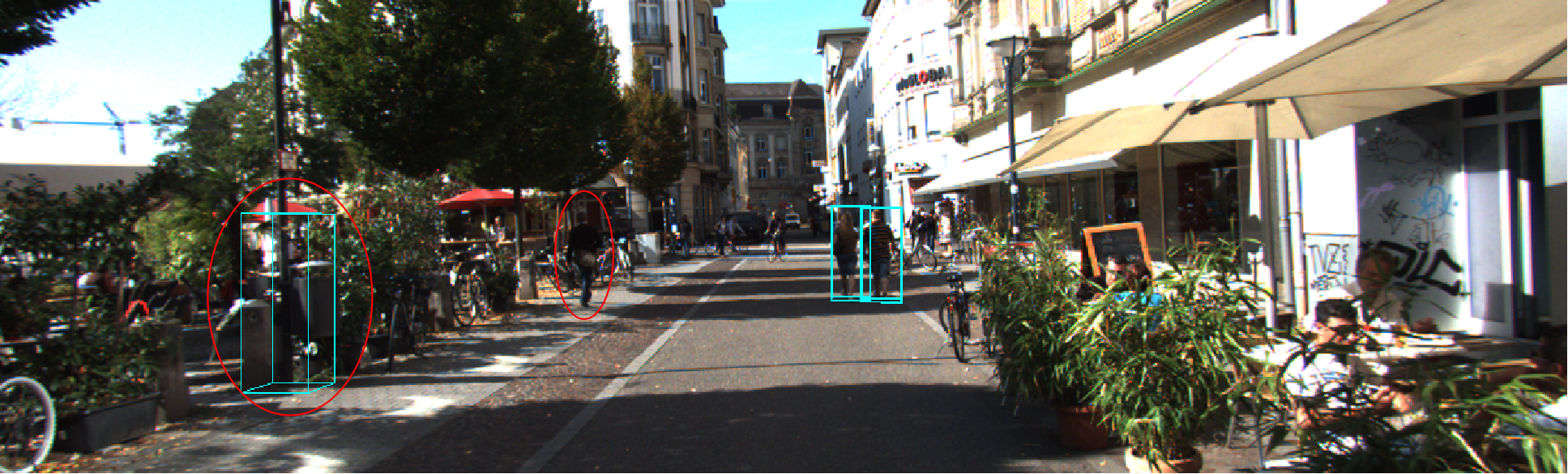}\vspace{8pt}
	\end{minipage}
	\begin{minipage}{0.495\linewidth}
		\centering
		\includegraphics[width=0.98\linewidth]{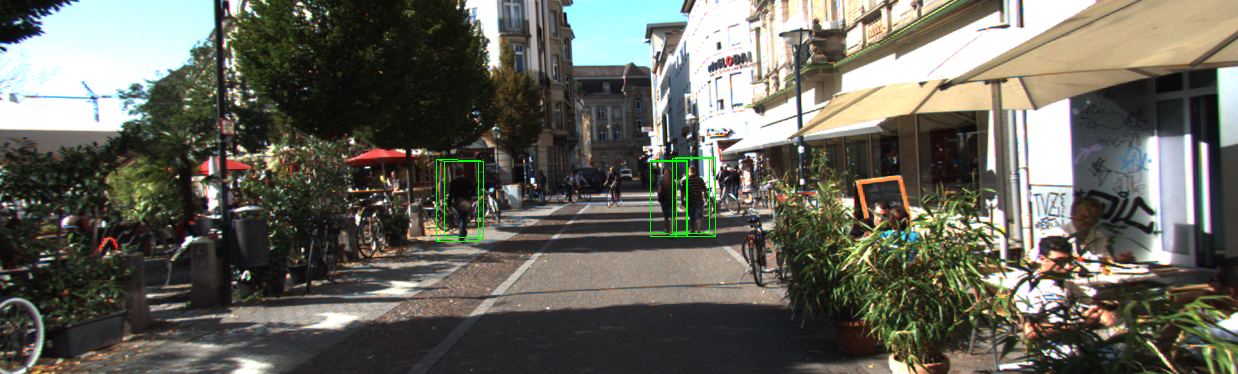}\vspace{8pt}
	\end{minipage}
    \begin{minipage}{0.495\linewidth}
		\centering
		\includegraphics[width=0.98\linewidth]{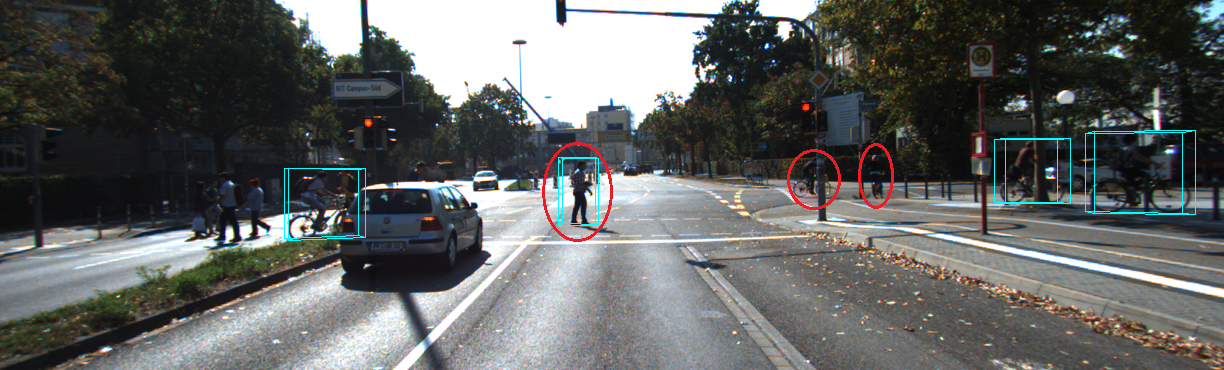}\vspace{8pt}
	\end{minipage}
	\begin{minipage}{0.495\linewidth}
		\centering
		\includegraphics[width=0.98\linewidth]{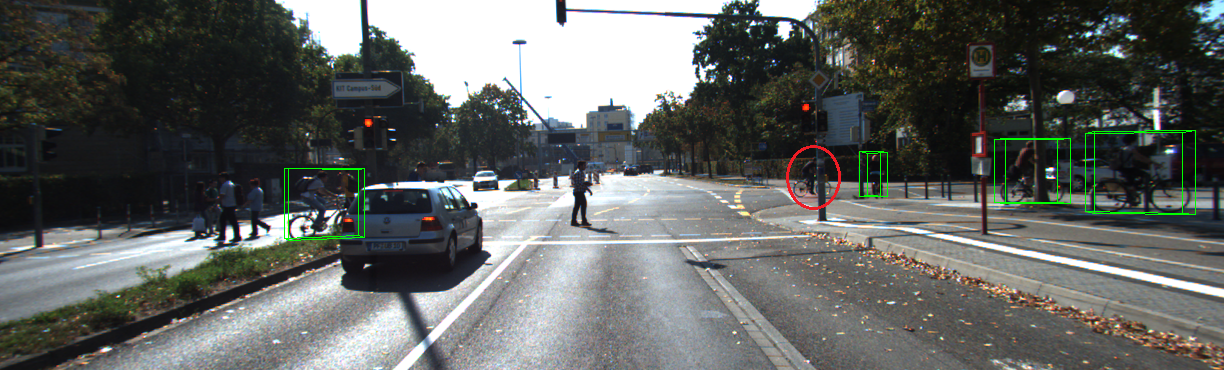}\vspace{8pt}
	\end{minipage}
	\caption{Four representative examples to visualize the detection results between MonoDETR (left) and SSD-MonoDETR (right), where the red circles indicate the missing objects.}
	\label{duibi}
\end{figure*}

\textbf{Quality evaluation on query points}:
We evaluate the quality of query points by the three approaches: MonoDETR, SSD-MonoDETR, and SSD-MonoDETR/WSM (SSD-MonoDETR without using WSM loss), and Table~\ref{ER} shows the comparison results, which are measured by two quantitative criteria: Position Precision, and Weighted Position Precision. Position precision is equal to the proportion of the number of key points falling inside objects to the total number of all the key points, while weighted position precision additionally multiplies the predicted attention weight of each key point to calculate position precision. Benefiting from SSDA, SSD-MonoDETR could better estimate the scales of key points and thus achieves better position precision as compared to MonoDETR. Without using WSM loss, SSD-MonoDETR/WSM suffers from a sharp performance drop, which proves the effectiveness of WSM loss in supervising the scale prediction. Moreover, we discover that the improvement in weighted position precision is enlarged as compared to position precision, which indicates that SSD-MonoDETR gives large attention weights to key points inside objects.\\
\indent Figure~\ref{point} shows an example to visualize the generated key points by MonoDETR and SSD-MonoDETR. Obviously, as compared to MonoDETR, the distribution of the predicted key points by SSD-MonoDETR is more concentrated on objects with larger weights. As a result, SSD-MonoDETR could better learn query features to support object detection.

\subsection{Evaluation of Weighted Scale Matching Loss}
In this experiment, we first study with different weights $\lambda_8$ of the Weighted Scale Matching (WSM) loss in Equation~\ref{all loss} and then evaluate the effectiveness of the penalty weight item $W(q)$ in Equation~\ref{Weight}.

\begin{table}[ht]
	\begin{center}
		\renewcommand{\arraystretch}{1.2}
  \caption{Performance trend with respect to different values of $\lambda_8$.}
		\setlength{\tabcolsep}{2.5pt}{
			\begin{tabular}{|m{2cm}<{\centering}|m{1.5cm}<{\centering}m{1.5cm}<{\centering}m{1.5cm}<{\centering}|}
				\hline
				$\lambda_8$ & Easy & Mod. & Hard \\
				\hline\hline
				0 & 27.53 & 19.54 & 16.30 \\
				0.1  & 28.85& 20.31 & 17.14\\
				0.2  & $\bm{29.53}$ & $\bm{21.96}$ & $\bm{18.20}$  \\
				0.3  & 29.06 & 20.57 & 17.67 \\
				0.4  & 28.04 & 19.67 & 17.03 \\
				0.5  & 26.33 & 18.91 & 15.84 \\
				\hline
		\end{tabular}}
		\label{ab1}
	\end{center}
\end{table}
Table \ref{ab1} demonstrates the performance trend with respect to different weights. When $\lambda_8$ increases from $0$ to $0.2$, the performance is significantly improved by about $2\%$ on all the samples. This is because the use of WSM loss supervises the scale prediction for query points, and thus could offer better query features for object detection. Moreover, the further increase of $\lambda_8$ would lead to a performance drop since a too-large weight on WSM loss would overshadow the utility of $L_{2D}$, $L_{3D}$ detection losses, yielding prediction errors.\\

\begin{table}[t!]
	\begin{center}
		\renewcommand{\arraystretch}{1.2}
  \caption{Performance comparison among the three weighting schemes for the WSM loss.}
		\setlength{\tabcolsep}{2.5pt}{
			\begin{tabular}{|m{2.5cm}<{\centering}|m{1.5cm}<{\centering}m{1.5cm}<{\centering}m{1.5cm}<{\centering}|}
				\hline
				Form & Easy & Mod. & Hard \\
				\hline\hline
				WSM loss-0 & 28.81 & 20.68 & 17.12 \\
				WSM loss-L1  & 29.12& 21.05 & 17.79\\
				WSM loss  & $\bm{29.53}$ & $\bm{21.96}$ & $\bm{18.20}$  \\
				\hline
		\end{tabular}}
		\label{wsm-loss}
	\end{center}
\end{table}

\indent To verify the effectiveness of the penalty weight item $W(q)$ in Equation \ref{Weight}, we compare our WSM loss with the following two settings: WSM loss-0 without using the weight $W(q)$ by setting $W(q)=1$, and WSM loss-L1 by setting $W(q)=log L(q)$ which indicates the weight of a query is proportional to its prediction error.  Table \ref{wsm-loss} shows the performance comparison results. Compared to WSM loss-0, WSM loss-L1 achieves better performance because it assigns larger training weights to error samples. Furthermore, our WSM loss considers the global-aware weighting correlations among all the queries in a training batch and thus could achieve the best performance.

\subsection{Qualitative Results}
Figure~\ref{duibi} shows six visualized examples with the detection results by MonoDETR and SSD-MonoDETR, where the first to fourth lines are the detection results of ``Car'', the fifth and sixth lines are the visualization of ``Pedestrian'' and ``Cyclist'', respectively.
We discover that several small and partially blocked cars are missed by MonoDETR, as MonoDETR is easy to lose relevant query points for these hard objects. When it comes to the ``Pedestrian'' and ``Cyclist'' categories, which always have relatively small scales, MonoDETR generates more severe instances of missed or false detection.
Comparatively, SSD-MonoDETR estimates the scale of an object query to generate more inner-the-object key points, as shown in Figure~\ref{point}, thus extracting more relevant local features, which offers robust query features to the detection heads to support accurate detection, especially on hard samples.

\section{Conclusions} \label{conclusions}
We propose SSD-MonoDETR to first introduce Supervised Scale-aware Deformable Attention (SSDA) for monocular 3D object detection. Different from the existing transformer-based methods, SSDA could estimate the scale of a query to better capture its receptive field to impose the scale awareness on key point prediction, yielding better query features for 3D attribute prediction.
Aside from this, SSDA adopts supervised learning with a WSM loss without extra labeling costs, which is more effective as compared to the unsupervised attention in transformers. Extensive experiments and analyses on KITTI have demonstrated the effectiveness of our approach. However, the proposed SSDA inevitably requires pre-setting different scales for different categories of objects, generating extra training costs on different categories.
Furthermore, there is space to improve the testing speed of our approach.
In the future, we intend to design a more flexible and generalized attention module and embed it into more transformer-based 3D object detection backbones for their further performance improvement in scene understanding.

\bibliographystyle{IEEEtran}
\bibliography{egbib}

\end{document}